\documentclass[pdflatex,sn-mathphys]{sn-jnl}

\jyear{2022}%

\usepackage{setspace} 

\theoremstyle{thmstyleone}%
\theoremstyle{thmstyletwo}%

\theoremstyle{thmstylethree}%

\begin{document}

\title[Cognitive multilayer networks
]{Cognitive modelling with multilayer networks: Insights, advancements and future challenges}




\author*[1]{\fnm{Massimo} \sur{Stella}}\email{massimo.stella@inbox.com}

\author[2,3]{\fnm{Salvatore} \sur{Citraro}}

\author[3]{\fnm{Giulio} \sur{Rossetti}}

\author[4]{\fnm{Daniele} \sur{Marinazzo}}

\author[5]{\fnm{Yoed N.} \sur{Kenett}}

\author[6]{\fnm{Michael S.} \sur{Vitevitch}}

\affil*[1]{\small \orgdiv{CogNosco Lab, Department of Computer Science, University of Exeter},  \orgaddress{ \city{Exeter}, \country{UK}}}

\affil[2]{\small \orgdiv{Department of Computer Science}, \orgname{University of Pisa}, \orgaddress{ \city{Pisa}, \country{Italy}}}

\affil[3]{\small \orgdiv{Institute of Information Science and Technologies}, \orgname{National Research Council}, \orgaddress{\city{Pisa}, \country{Italy}}}

\affil[4]{\small \orgdiv{Faculty of Psychology and Educational Sciences, Department of Data Analysis}, \orgname{University of Ghent}, \orgaddress{\city{Ghent},  \country{Belgium}}}

\affil[5]{\small \orgdiv{Cognitive Complexity Lab, Faculty of Industrial Engineering and Management}, \orgname{Technion, Israel Institute of Technology}, \orgaddress{ \country{Israel}}}

\affil[6]{\small \orgdiv{Department of Psychology}, \orgname{University of Kansas}, \orgaddress{ \city{Lawrence}, \state{Kansas}, \country{USA}}}


\abstract{The mental lexicon is a complex cognitive system representing information about the words/concepts that one knows. Decades of psychological experiments have shown that conceptual associations across multiple, interactive cognitive levels can greatly influence word acquisition, storage, and processing. How can semantic, phonological, syntactic, and other types of conceptual associations be mapped within a coherent mathematical framework to study how the mental lexicon works? We here review cognitive multilayer networks as a promising quantitative and interpretative framework for investigating the mental lexicon. Cognitive multilayer networks can map multiple types of information at once, thus capturing how different layers of associations might co-exist within the mental lexicon and influence cognitive processing. This review starts with a gentle introduction to the structure and formalism of multilayer networks. We then discuss quantitative mechanisms of psychological phenomena that could not be observed in single-layer networks and were only unveiled by combining multiple layers of the lexicon: (i) multiplex viability highlights language kernels and facilitative effects of knowledge processing in healthy and clinical populations; (ii) multilayer community detection enables contextual meaning reconstruction depending on psycholinguistic features; (iii) layer analysis can mediate latent interactions of mediation, suppression and facilitation for lexical access. By outlining novel quantitative perspectives where multilayer networks can shed light on cognitive knowledge representations, also in next-generation brain/mind models, we discuss key limitations and promising directions for cutting-edge future research.}




\maketitle

\section{Introduction}\label{sec1}

The mental lexicon is a complex system where knowledge of the words and concepts one knows can be represented as units that are combined and associated across multiple levels \cite{aitchison2012words}. For example, phonemes combine to form words, words combined in sentences express ideas, and sentences in narratives give rise to stories \cite{vitevitch2020network,litovsky2022semantic}. Focusing on the level of units of words (which provide meaning even in isolation), deeper knowledge can be expressed by linking together units that are associated in some way. Words can be associated in many ways \cite{fillmore2006frame,zock2020comparison,aitchison2012words}. For example, words may share meaning \cite{steyvers2005large}, sound similar \cite{vitevitch2008can}, be syntactically related \cite{semeraro2022emotional}, bring each other to mind \cite{de2013better}, represent objects with similar semantic or visual features \cite{kennington2015simple}, be written similarly \cite{siew2019phonographic} or evoke the same set of emotions and affective states \cite{mohammad2013crowdsourcing}. These are only some of the many ways in which words can be associated \cite{aitchison2012words,vitevitch2020network,stella2020multiplex} and give structure to the knowledge that one has that can be expressed through language. Decades of research in psycholinguistics and cognitive science have examined how the words and concepts in the mental lexicon are acquired, stored, processed, and retrieved \cite{collins1975spreading,aitchison2012words,zock2020comparison}. Importantly, it has been shown that the structure and organisation of the words and concepts associated in some way in the mental lexicon influence a wide variety of linguistic and cognitive phenomena, such as word confusability \cite{vitevitch2020network}, picture naming \cite{castro2019multiplex,castro2020contributions,castro2020quantifying}, and memory recall patterns for both neutral \cite{de2013better,de2019small,kenett2017semantic,montefinese2015semantic} and emotional information  \cite{fatima2021dasentimental,martinelli2021cognitive}. The structure and organisation of the words and concepts associated in some way in the mental lexicon can be influenced by various factors, including psychedelic drugs \cite{rastelli2022simulated}, and how creative \cite{kenett2018flexibility}, expert \cite{koponen2021systemic} or curious \cite{zurn2018curiosity} an individual is. All these findings converge on one point: Understanding the structure and organisation of knowledge in the mental lexicon is important for shedding light on a number of phenomena. Understanding the structure and organisation of knowledge in the mental lexicon requires a framework that is quantitative \cite{siew2019cognitive}, interpretable \cite{rudin2019stop} and human-centric \cite{bryson2019society}. This framework must: (i) be capable of producing inferences and comparable measurements regulated by mathematical equations and theoretical models \cite{castro2020contributions,zemla2020snafu} (quantitative); (ii) map results to outputs through an internal representation of knowledge available to researchers, unlike most black-box machine learning knowledge models \cite{fatima2021dasentimental} (interpretable); and (iii) be grounded in psychological theory and large-scale datasets in order to account for the complex nuances of human psychology rather than make abstract inferences that are of little value to psychologists \cite{zaharchuk2021multilayer,wulff2022using}. An artificial intelligence that categorises individuals using binary labels like “aphasic” or “healthy” without identifying the severity of their language impairments, nor considers their ability to acquire, retain, and produce new knowledge would not be human-centric \cite{castro2020quantifying}).

In the present review, we advance the idea of using multilayer networks to model and understand the structure and organisation of knowledge in the mental lexicon. We discuss recent work from multiple fields to show how multilayer networks are a quantitative, interpretable, and human-centric framework that can connect several disparate disciplines interested in modelling knowledge. Multilayer networks are a cutting-edge approach to explore how knowledge is processed simultaneously across multiple levels. We outline 3 recent research developments where the ability to combine different layers of associative knowledge highlights phenomena that would be otherwise lost in single-layer network analyses or through other modelling approaches like word embeddings \cite{rudin2019stop}. We discuss key limitations of this framework and review potential approaches for future research in cognitive modelling \cite{kenett2016structure,vitevitch2021cognitive} and cognitive neuroscience \cite{poeppel2022brainmind,zaharchuk2021multilayer}. Combining evidence from fields as diverse yet interconnected as cognitive psychology, complexity science, and computer science, our review identifies concrete innovative ways in which multilayer networks can advance our understanding of cognition.

\section{Evidence for the multilayered nature of the mental lexicon}

Despite the name, the mental lexicon is not a simple dictionary \cite{vitevitch2020network,zock2020comparison,pirrelli2020word,doczi2019overview}. Concepts in the mental lexicon are not recalled in alphabetical order and the recollection of an item is not independent of other concepts associated with it \cite{kumar2020distant,hills2022mind}. Aitchison \cite{aitchison2012words} used the London tube as a metaphor for the mental lexicon, where stations represent linguistic units and are connected according to a layout of channels of different lengths. This analogy resonates with the concept of a complex network, although the exact specification of structure, function, and dynamics in the mental lexicon is more sophisticated \cite{doczi2019overview}. Even though the mental lexicon might not be a network itself, some of its associative features might be accurately modelled by network science \cite{hills2022mind}.

Many research findings indicate that information represented in the mental lexicon is inherently multi-layered: Phonological, semantic, and syntactic aspects of language can simultaneously interact and influence language retrieval and processing \cite{zock2020comparison,dell1997lexical,castro2020contributions}. In healthy populations, the interaction of multiple types of linguistic interactions in the mental lexicon is highlighted by the phenomenon of the tip of the tongue \cite{zock2020comparison}, where an individual is aware of the semantic features of a word but cannot produce it. This tip-of-the-tongue state is characterised by a failure to retrieve phonological information, whereas semantic activation seems to be intact \cite{pirrelli2020word,brown1991review}. Another example of faulty retrieval is known as a malapropism \cite{fay1977malapropisms,pirrelli2020word} where a similar sounding word is retrieved for another semantically appropriate one (e.g. “dancing a flamingo” instead of “dancing a flamenco”). The faulty interaction between semantic and phonological information of words can also explain the increase of mixed errors in people with aphasia in a picture naming task \cite{dell1997lexical}. 

Evidence for the multilayered nature of the mental lexicon comes also from facilitative effects in word production like priming \cite{aitchison2012words,kenett2017semantic,kumar2020distant}, i.e. when lexical retrieval is facilitated by cues related to target words. Morphological content (e.g. “dog” containing phonemes \textbackslash d\textbackslash, \textbackslash o\textbackslash  and \textbackslash g\textbackslash), synonym similarities (e.g. “character” and “nature” being synonyms), and syntactic relationships (e.g. being a certain part of speech) were found to facilitate lexical retrieval through priming indicating the simultaneous interplay between phonological, semantic, and syntactic layers of the mental lexicon \cite{bock1996language,doczi2019overview}. These findings motivated the formulation of the so-called cognitive linguistic theory \cite{de2013dynamic}, of serial lexical access \cite{dell1997lexical} and of cobweb theory \cite{aitchison2012words}, which all argue that language production depends on a network of interacting layers of the mental lexicon, including individual phonemes, word meaning, and sentence structuring. Given the interaction of various types of information in the lexicon, the framework of multilayer networks becomes a natural way of analysing the structural and dynamical complexity of the mental lexicon. 

\section{From single-layer to multilayer networks as models of cognition}

Complex networks represent the structure of pairwise connections between interacting entities \cite{newman2018networks}. Connections are usually called links or edges, and the interacting entities are usually called nodes or vertices. A single-layer complex network represents only one type of relationship between nodes. Instead, both “multiplex” and “multilayer” networks include multiple types of relationships between nodes \cite{bianconi2018multilayer}. For the sake of an easier visualisation and to fully characterise the mathematics of such multilayer/multiplex networks, usually, nodes are organised in sub-groups called network layers \cite{de2015structural,santoro2020algorithmic}, which identify specific aspects of the pairwise interactions between individual nodes. A single network layer is composed of a specific type of interaction between nodes \cite{stella2017multiplex}. Links connecting nodes from the same network layer are called intra-layer links. Links connecting any two nodes from different network layers are called inter-layer links \cite{bianconi2018multilayer}.

Whereas single-layer networks can account only for one type of associative links, e.g. syntactic relationships \cite{correa2020semantic}, the main advantage of multilayer networks is the ability to combine multiple types of associations within a single model \cite{boccaletti2014structure,de2013mathematical,kivela2014multilayer,battiston2020networks}. The presence of multiple aspects or layers of associations can give rise to phenomena greatly different from single-layer networks, such as the presence of feedback loops across layers \cite{kivela2014multilayer,stella2018ecological}, changes in the centrality of individual nodes when multiple interactions are simultaneously present \cite{bianconi2018multilayer}, or the emergence of patterns of connectedness undetectable in the individual layers \cite{battiston2020networks,battiston2014structural}.

Both “multilayer” and “multiplex” networks have multiple network layers and represent multiple types of interactions, but these two terms are not synonyms \cite{bianconi2018multilayer,de2013mathematical,kivela2014multilayer}. Multilayer networks represent a more general category of complex networks, whereas multiplex networks are a more specific network model because they feature \textit{node alignment} \cite{battiston2020networks}. Node alignment means that the same set of nodes are found in every network layer, with the same node being connected across layers. The presence of explicitly weighted intra-layer connections characterises full multiplex networks. Although Collins and Loftus \cite{collins1975spreading} discussed the idea of multilayer phonological/syntactic networks of conceptual associations in the 1970's, Cong and Liu \cite{cong2014approaching} were the first to implement a multilayer representation of language representing syntactic and phonological relationships between Mandarin words. Importantly, Martinčić-Ipšić and colleagues \cite{martinvcic2016multilayer} extended the multilayer formalism to Croatian and English, highlighting several structural similarities between the two languages, which differ in terms of the allocation of syllables across words. Without node alignment, multilayer networks can feature different sets of nodes across layers, see Figure 1 (left). One layer might feature phonemes, linked to a layer of words by several interlayer links expressing how phonemes occur in words. Words might also be linked to another layer expressing their semantic features, the latter being connected by intra-layer links expressing antonyms. Featuring different nodes across different layers makes the mathematics describing multilayer networks considerably more advanced than the mathematics behind multiplex networks \cite{bianconi2018multilayer,boccaletti2014structure}. Whereas a multiplex network can be described with matrices, multilayer networks require the use of tensors to represent them (for more details, see \cite{de2013mathematical} and \cite{kivela2014multilayer}).

Multilayer networks featuring node alignment are called multiplex networks \cite{battiston2020networks}. Node alignment requires the same set of nodes to be replicated across all layers of a multiplex network, see also Figure 1 (right), where alignment is represented by dashed lines. Even if a node is disconnected on one layer of a multiplex network, but highly connected on another layer the node has to appear in both layers. All the replicas of a given node represent different aspects of the same entity, e.g. a phonological word form engaging in phonological relationships on a given layer and a lexical representation of the same word involved in conceptual associations on another layer \cite{stella2018multiplex,stella2020multiplex}. Replica nodes identify the so-called physical node, e.g. in the above example the phonological form and the lexical representation are replicas identifying the same word/physical node \cite{kivela2014multilayer,de2013mathematical,battiston2014structural}. 
Figure 1 provides an example of a multilayer language network, analogous to pioneering work by Cong and Liu \cite{cong2014approaching}, and a multiplex lexical network, analogous to pioneering work by Stella and colleagues \cite{stella2017multiplex}. The layered structure of multilayer and multiplex networks enables the possibility to include both semantic and phonological features of linguistic units. For instance, Figure 1 (left) identifies semantic features and phoneme occurrences of words (as in \cite{martinvcic2016multilayer}), whereas Figure 1 (right) maps semantic overlap, phonological similarities and free associations between words (as in \cite{stella2018multiplex}). Notice that in multiplex networks it is only the type of interactions among nodes that changes across layers, whereas multilayer networks allow for a more flexible structure with the possibility of different types of nodes across layers. Notice also that in the absence of explicitly weighted intra-layer links, multiplex networks become edge-coloured graphs \cite{battiston2020networks,semeraro2022emotional}, where connections of different types are coloured differently. The presence of multiple layers/interactions alters drastically the connectivity of words compared to their single-layer layouts \cite{cong2014approaching,martinvcic2016multilayer}. For instance, in Figure 1 (right), "mat" is disconnected on the free association layer but connected to "cat" on the phonological layer. Multiple layers also give rise to edge overlap across different aspects of associative knowledge, e.g. "kitty" and "cat" share a connection on both the free association and the semantic overlap layers, an overlap that cannot be measured when layers are considered as separate, single components \cite{de2022multilayer}. In the following, we review how multilayer/multiplex networks can identify and quantify cognitive patterns that would go undetected using standard single-layer networks.

\begin{figure}
    \centering
    \includegraphics[width=12cm]{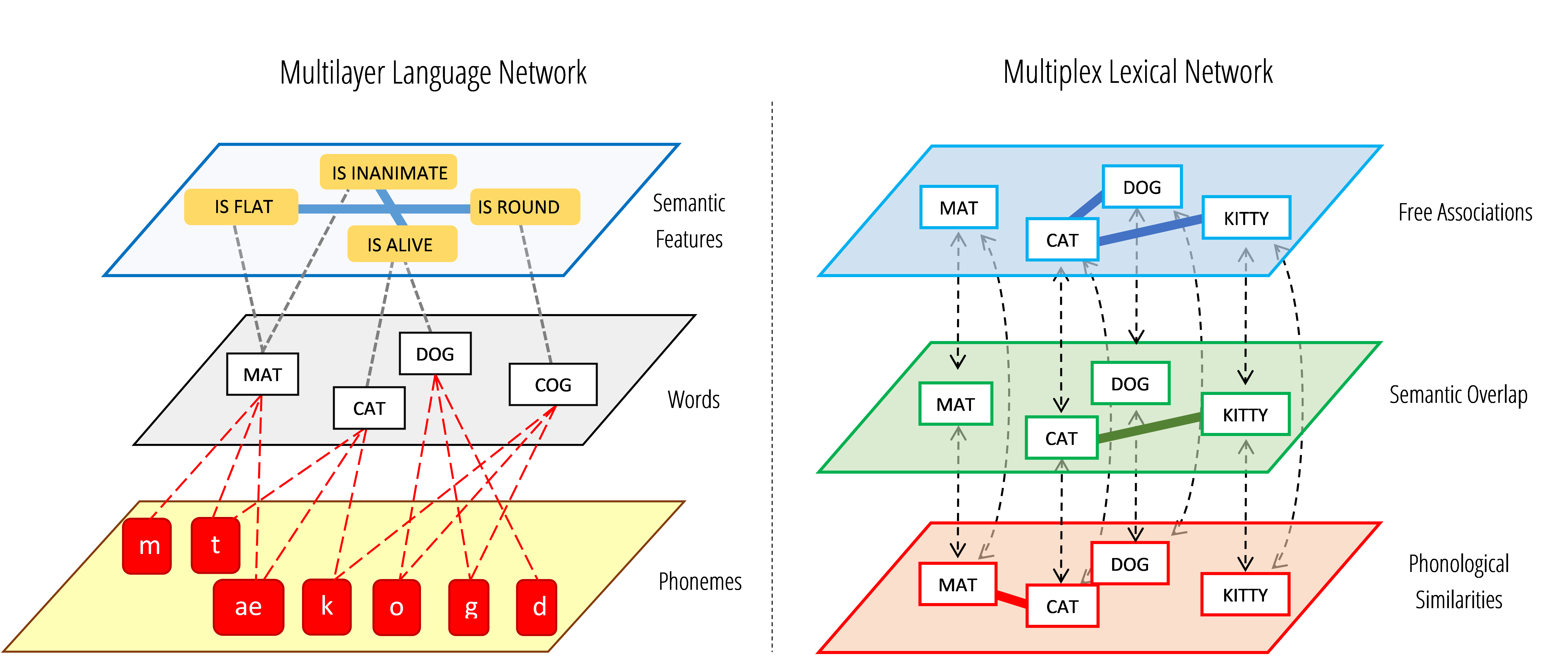}
    \caption{Examples of a multilayer language network (left,  \cite{cong2014approaching}) and of a multiplex network (right,  \cite{stella2017multiplex}). If inter-layer connections were all unweighted, the multiplex network would be an edge-coloured graph with three colors (free associations, semantic overlap and phonological similarities). The multilayer language network maps semantic features and phoneme occurrence in words. Node alignment, the replication of the same set of nodes across layers, is a feature characterising multiplex networks.}
    \label{fig:1}
\end{figure}


\section{Cognitive patterns highlighted by multilayer networks but invisible to single-layer networks}

Multilayer networks take into account more than one type of relationship among nodes at once, thus giving rise to more complex structures and phenomena that cannot be observed in single-layer networks. This section reviews key ways in which multilayer networks have been shown to differ from single-layer networks. We discuss these quantitative differences in relation to the relevant psychology literature, and consider the overall benefits, limitations, and roads for future research of multilayer networks as models of cognition.

\subsection{Multiplex viability highlights language kernels in the mental lexicon invisible to single-layer networks}

As in single-layer networks \cite{siew2019cognitive}, the collection of all intra-layer and inter-layer links connecting nodes $i$ and $j$ also represents a path in multilayer networks \cite{bianconi2018multilayer}. For instance, in Figure 1 (right) there is a multilayer path connecting “mat” and “kitty” through the intra-layer link “mat (phonological)” - “cat (phonological)”, the interlayer link “cat (phonological)” - “cat (semantic overlap)” and the intra-layer connection “cat (semantic overlap)” - “kitty (semantic overlap)”. In contrast to describing a path between nodes in a single-layer network, in multilayer paths it is necessary to specify the layers of nodes to distinguish intra- and inter-layer links. In edge-coloured multiplex networks, this distinction is not necessary but links of different colours must be kept separate \cite{vukic2020structural,de2022multilayer}. Assuming specific weights for inter- and intra-layer links enables the definition of the shortest path length, i.e. the minimum total weight or number of links necessary for traversing a path between any two nodes. In edge-coloured graphs, the shortest path length (also called network distance or geodesic network distance) considers only intra-layer links and can be defined as the smallest number of links of any colour connecting any two nodes \cite{fatima2021dasentimental}. Without considering explicit inter-layer connections, the network distance between “mat” and “kitty” would be 2 in Figure 1 (left). Ultimately, the possibility of “jumping” across layers enhances the connectivity of multilayer networks. Multilayer shortest path lengths were shown to significantly predict cognitive phenomena like early word acquisition \cite{stella2017multiplex,stella2018distance,stella2019modelling,citraro2022feature}, semantic relatedness \cite{levy2021unveiling} and picture naming production in people with aphasia \cite{castro2020quantifying,castro2019multiplex}. In all of these studies, multiplex network distances achieved better model performances than their single-layer counterparts, thus providing quantitative, converging evidence that the ability to transition between layers enabled by multilayer networks is crucial to modelling the interactive aspects of the mental lexicon.

Interactions between different aspects of the lexicon might be modelled through explicit \cite{levy2021unveiling,stella2018cohort} or latent \cite{marinazzo2022information} patterns of connectivity between nodes in different layers. Whereas we discuss latent patterns in Subsection 4.3, here we focus on explicit interactions arising from connectivity patterns. Transitioning between multiple layers gives rise to multiple ways of defining connectedness in multilayer networks, see Figure 2. In single-layer networks, two nodes are connected if there exists a path between them \cite{newman2018networks}. Analogously, several works on multilayer and multiplex networks defined connectedness as depending on the existence of a multilayer path between any two nodes \cite{boccaletti2014structure,de2013mathematical,battiston2014structural}. The largest connected component can then be defined as the largest set of nodes connected by at least one multilayer path \cite{stella2018cohort,stella2018multiplex,battiston2020networks,bianconi2018multilayer}. This definition can be modified in the presence of explicitly defined inter-layer links \cite{bianconi2018multilayer}.  Importantly, the presence of several layers can give rise to additional definitions of connectivity that differ from their single-layer counterparts \cite{newman2018networks}, thus giving rise to phenomena unobserved in single-layer networks, such as the identification of language kernels \cite{cancho2001small,aitchison2012words}.

\begin{figure}
    \centering
    \includegraphics[width=9.5cm]{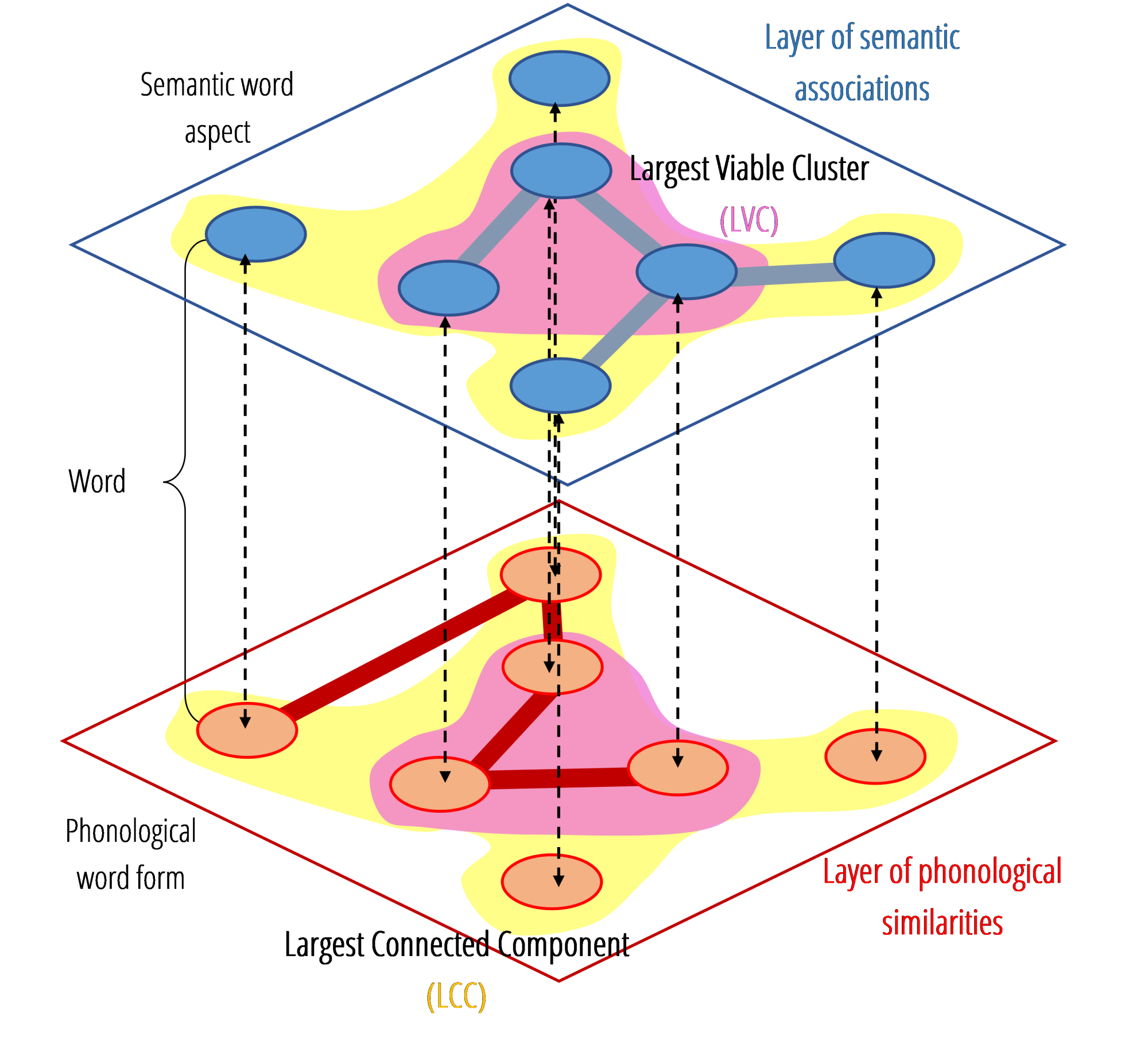}
    \caption{In multiplex lexical networks, connectivity can be defined either in terms of connected components or in terms of viable clusters. The largest connected component (LCC, orange) is the largest set of nodes connected by at least one path with links of any colour. The largest viable cluster (LVC, pink) is the largest set of nodes connected simultaneously on all layers by at least one path with links of the same colour. The LVC is a more restrictive version of the LCC but the two coincide in single-layer networks.}
    \label{fig:2}
\end{figure}

The above multilayer definition of connectedness \cite{bianconi2018multilayer} exploits links present in either one layer or others. This combinatorial “OR” approach is different from requiring the presence of connected paths across one layer and others. Other works \cite{baxter2014avalanches,bianconi2018multilayer,baxter2021weak} considered \textit{viability} as a definition of connectedness based on an “AND” logic: In a multilayer network, two nodes are viably connected if there exist intra-layer paths connecting those two nodes on every single layer. Intra-layer paths are confined to a single path, e.g. paths using links of only one colour. A largest viable component (LVC) is the largest set of viably connected nodes \cite{baxter2021weak,stella2020multiplex}. As highlighted in Figure 2, the requirement of intra-layer connectedness across all layers (i.e. viability) is considerably more restrictive than the above definition of connectedness. This distinction naturally leads to the question of understanding whether connected components and viable components might differ in their structure when modelling cognitive aspects of the mental lexicon\cite{stella2018multiplex}, e.g. contain different sets of concepts. Note that in single-layer networks, the largest viable cluster and the largest connected component (LCC) would be the same \cite{baxter2014avalanches}. However, the LVC and the LCC would differ on multilayer networks made of different layers, potentially giving rise to phenomena unexpected in single-layer cognitive networks \cite{siew2019cognitive}.

\subsubsection{The LVC corresponds to a spurt in language learning}

The first quantitative evidence characterising the cognitive relevance of LVCs was \cite{stella2018multiplex}. The authors identified an LVC of 1000 words in a representation of the mental lexicon with an LCC of 8000 English words, connected across 4 semantic/syntactic/phonological layers. Bringing together multiple psycholinguistic datasets, the authors simulated the growth of the LVC over time according to normative age of acquisition norms, imitating the order in which most native English speakers acquire concepts over time. Whereas the LCC grew smoothly over time, the LVC appeared with a sudden, discontinuous phase transition (called also explosive \cite{baxter2021weak}) around age 7-8 yrs, a critical age for the development of reading and reasoning skills in typically developing children \cite{piaget2003part,frith2017beneath}. The LVC was also found to be rich in shorter and higher frequency/polysemy/concreteness words (compared to the LCC). When partitioning words as inside/outside of the LVC, each multiplex layer exhibited a core-periphery structure \cite{newman2018networks}, with the LVC representing a network core, i.e. a set of tightly linked high-degree nodes connecting more peripheral low-degree nodes. Furthermore, when removing words in the LVC from the multiplex lexical network, the average network distance between words increased considerably more than removing words outside of the LVC but with a matched degree. Since distance in cognitive networks refers to conceptual similarities \cite{siew2019cognitive,fatima2021dasentimental} and shorter distance corresponds to quicker conceptual processing (cf. \cite{kenett2017semantic,kumar2020distant,vitevitch2020network}), this pattern indicates a beneficial role played by nodes in the LVC in providing shortcuts of conceptual associations between other concepts. Noticeably no LVC was found when the phonological layer was excluded from the analysis, suggesting that phonological associations are key to identifying the LVC. Given all of these characteristics of the LVC, Stella and colleagues suggested that the LVC was a language kernel in the mental lexicon \cite{cancho2001small}. That is, the LVC is a sample of highly frequent yet simple words that are prominent (e.g. well connected) in language and whose availability enables communicative advantages \cite{aitchison2012words,cancho2001small,gruenenfelder2009lexical}.

\subsubsection{The LVC identifies changes in mental navigation related to variation in creativity}

Subsequent research using the LVC recently demonstrated its ability to classify low- and high- creative individuals \cite{stella2019viability}. Stella and Kenett analysed performance in a semantic fluency task, as an operationalisation of a mental navigation task that operates over memory when searching internally \cite{todd2020foraging,rastelli2022simulated,abbott2015random}. In that task, participants are required to generate as many category members as possible, in a given amount of time. Computational methods allow examining how people search through their memory \cite{hills2015exploration,hills2012optimal,abbott2015random}, tracing the paths they traverse over representations of their mental lexicon \cite{hills2022mind}. Often, this task is based on the animal category \cite{hills2012optimal,goni2011semantic}, i.e. name as many animals as possible in 2 minutes. Specifically, Stella and Kenett re-analysed animal fluency data generated by low- and high- creative individuals, collected by Kenett and colleagues \cite{kenett2016structure}.

\begin{figure}
    \centering
    \includegraphics[width=12cm]{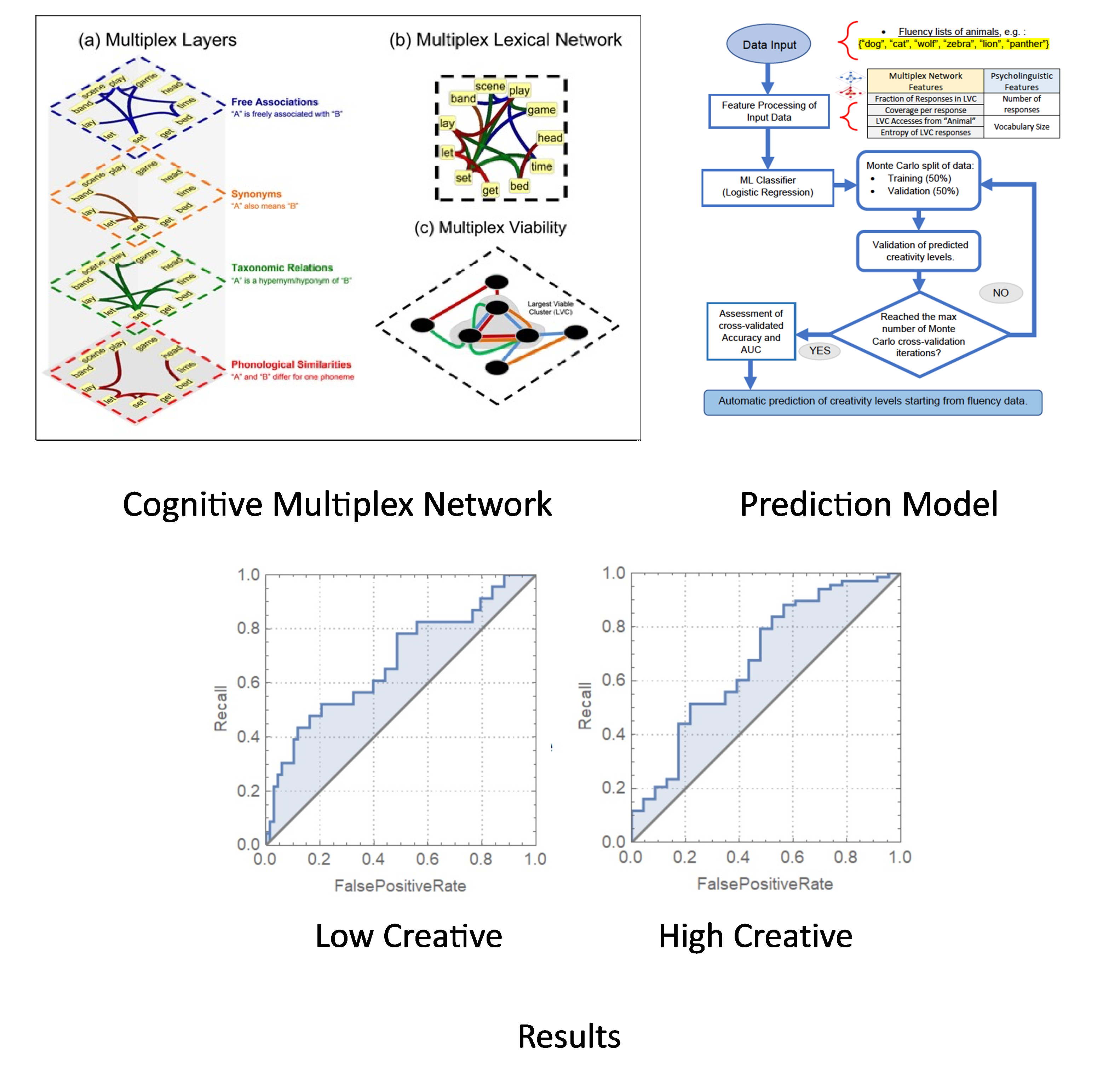}
    \caption{Illustration of the process taken by Stella and Kenett in their study \cite{stella2019viability}. First, the authors constructed a cognitive multiplex network (top left). next, the authors quantitatively examined participants’ semantic fluency responses as a mental navigation process over the cognitive multiplex network (top right). Participants’ measures of how they explore the cognitive multiplex network when searching for animal names was then used to build a machine learning model to classify participants in low or high creative individuals. This predictive model was highly successful in classifying participants into their correct group of low- and high-creative individuals (bottom).}
    \label{fig:fig6}
\end{figure}

To achieve creativity classification, the authors used a multiplex network to represent lexical memory over the same four layers used in prior work \cite{stella2018cohort} but extended to 16000 English words and exhibiting an LVC with over 1.1k words, see Figure \ref{fig:fig6}. Stella and Kenett examined computationally the way people exploited their memory, and classified participants into low- and high- creative individuals, based on the way they "walked" on the multiplex network, through the LVC. In other words, the authors analysed participants' measures of navigating over the cognitive multiplex network focusing on the LVC. The authors found that the low- and high- creative individuals differed in several cognitive multiplex measures, largely focusing on how they rely in their performance on the LVC, and in the number of responses they are able to generate. Individuals with lower creativity accessed the LVC considerably more than those with higher creativity, suggesting a beneficial role for the LVC to support recall in people unable to employ other cognitive strategies to achieve higher levels of creativity. Such distinctive patterns were measured through network access, distance, and entropy, and became a set of features in an artificial intelligence (AI) model that was trained to categorise high/low creativity levels from network measures of LVC access. In a leave-one-out cross-validation, the AI achieved an accuracy of $65 \pm 1\%$ (compared to a baseline of 50\% for random guessing), indicating that multiplex viability may influence conceptual recall and high-level cognitive strategies related to creativity \cite{kenett2018flexibility,beaty2022semantic}. 

Overall, the work of Stella and Kenett is significant in two ways. First, it demonstrates how a viable component in a multiplex network can be used to examine complex cognitive processes, such as mental navigation operationalised via a semantic fluency task. Second, the LVC can be used to extract features and construct machine learning predictive models, successfully predicting how people vary on complex cognitive capacities, such as creativity. Such evidence opens up the door to additional, future studies in representing complex cognitive processes and capacities.


\subsubsection{The LVC supports correct picture naming in people with aphasia}

Another example illustrating an advantage for words in the LVC comes from picture naming in people with aphasia. In a picture naming task individuals are visually presented a line drawing or photograph and asked to name the object that is depicted, a task formalised in the Philadelphia Naming Test (PNT), a 175-item picture naming test developed in the Language and Aphasia Lab of MRRI \cite{roach1996philadelphia}. Aphasia describes a spectrum of language disorders, impacting word processing, understanding and production \cite{dell1997lexical}. Castro and colleagues \cite{castro2020quantifying,castro2019multiplex} investigated picture naming through a multiplex lexical network with 8000 words linked by free associations, hypernyms/hyponyms, phonological similarities, and synonyms. The authors found that multiplex distance was important for predicting not only the rate of correct picture naming \cite{castro2019multiplex} but also the types of mistakes made by people with aphasia \cite{castro2020quantifying}. Higher predictive power in the first (regression) and second (multinomial regression) tasks were achieved when multilayer distances were used, rather than considering network layers in isolation. Building on those findings, Stella \cite{stella2020multiplex} discovered that items in the LVC were named correctly with rates at least 30\% higher than items found outside of the LVC. Further, through network attacks accounting for frequency, degree, and word length effects, the author found that the probability of correct production in PNT could efficiently identify words within the LVC (compared to random guessing \cite{stella2020multiplex}). Together these results suggest that words in the LVC might benefit from enhanced lexical retrieval mechanisms that in clinical populations leads to more accurate production of words in the LVC (as measured by the PNT \cite{stella2020multiplex}), and in healthy populations supports recall in individuals with lower creativity levels \cite{stella2019viability}. 

To sum up, in multilayer representations of the mental lexicon, viability can identify a language kernel that has interesting features for cognitive processing. This kernel emerges from the interactive nature of semantic and phonological associations \cite{stella2018multiplex,levy2021unveiling}, and facilitates cognitive processing in both healthy and clinical populations. Importantly, it is not possible to identify such a kernel in single-layer networks that model only part of the mental lexicon \cite{stella2017multiplex,stella2019modelling,stella2019viability}, highlighting the importance of using the multilayered network approach to shed light on cognitive representation and processing. Future research should further investigate clusters like the LVC and identify new ones, potentially arising from other types of relationships among words or by including other pieces of lexical information in the network \cite{citraro2022feature}. The clusters that are discovered in multilayer networks could lead to novel insights and provide quantitative ways to examine cognitive processing \cite{vitevitch2021cognitive}, creativity  \cite{beaty2022semantic}, cognitive functions in altered states of conscience \cite{rastelli2022simulated} and language acquisition \cite{citraro2022feature}, among other processes relevant to cognitive network science \cite{castro2020contributions,siew2019cognitive}.

\subsection{Community detection in multilayer networks highlights shortcuts between semantic themes}

 A community is a group of nodes more closely or tightly connected to each other than with nodes belonging to other groups \cite{blondel2008fast}. Community detection is the task of decomposing a network into well-connected and well-separated groups of nodes, and it is one of the most challenging problems in complex network analysis \cite{fortunato2016community}, in part because of the different topological criteria adopted to define a "community" \cite{newman2018networks}. Community detection algorithms, used to identify communities in a network, can be classified according to the way they approach the community detection task \cite{blondel2008fast,palla2005uncovering,girvan2002community}. The most common algorithmic approaches used to partition single-layer networks fall into two classes, adopting either (i) the well-known modularity-based optimisation scheme (see for details \cite{blondel2008fast}) or (ii) the concept of k-cliques - subgraphs where all nodes are adjacent to each other and have degree $k$ - to extract sets of overlapping communities \cite{palla2005uncovering}. Both community detection approaches have found clusters of words sharing similar linguistic traits in single-layer networks, like shared sequences of phonemes in phonological networks \cite{siew2013community}, or concepts falling in the same semantic field (found in a free association network \cite{palla2005uncovering} and in a syntactic network \cite{gerow2014modular}).

The task of community detection is more complicated in multilayer networks, because the community detection algorithm must consider the different types of relationships occurring in different layers at the same time \cite{magnani2021community}. Multilayer community detection algorithms are classified according to the strategy chosen to handle the presence of multiple layers:
\begin{itemize}
    \item Flattening methods reduce all layers into a single one, making the structure suitable for classic community detection \cite{berlingerio2011foundations}. This approach was used by Vukić and colleagues, who identified different semantic fields of "database" via community detection in a multilayer network with factual, conceptual, procedural, and metacognitive connections between concepts \cite{vukic2020structural}.
    \item Layer by layer methods process layers independently before merging the final list of communities through consensus, \cite{tagarelli2017ensemble}.
    \item Multilayer methods act directly on the multilayer structure, finding communities by transitioning across the layers. In this class of methods, we find extensions of the modularity-based approaches in single-layer networks to multilayer networks. Examples of this approach include GLouvain \cite{mucha2010community} or the multilayer extension \cite{edler2017mapping} of the Infomap algorithm \cite{rosvall2008maps}.
\end{itemize}

\begin{figure}
    \centering
    \includegraphics[width=12cm]{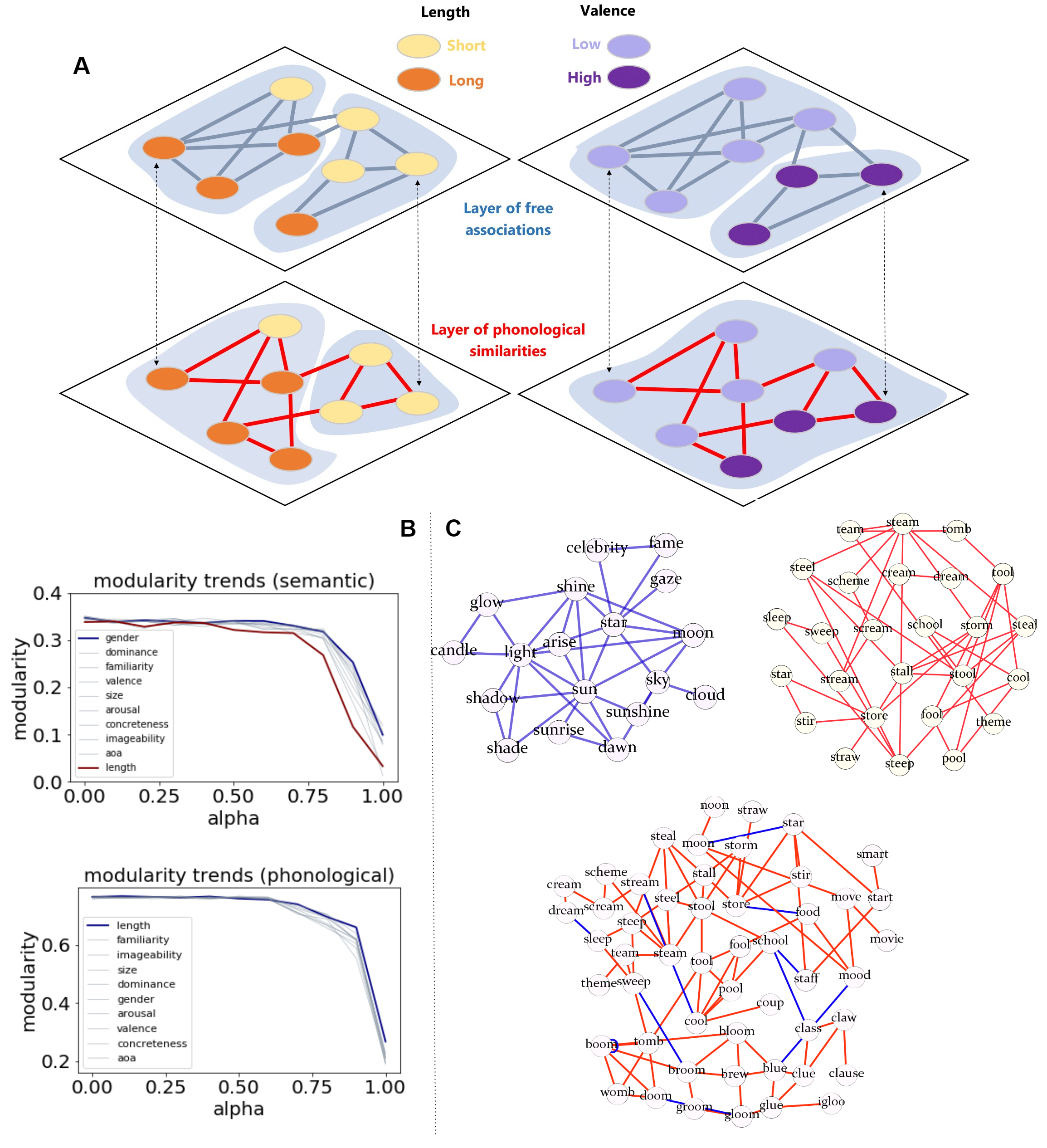}
    \caption{(A): Attribute-aware community detection in a toy multilayer lexical network: a toy partition for word length (left) and valence (right). Results from \cite{citraro2020identifying} (B-C): different attributes correlate with different layers (B); a toy example of matched communities on the semantic (blue links), phonological (red links), and multiplex (mixed links) lexical structure for the arousal property (C).}
    \label{fig:fig3}
\end{figure}

Choosing the most suitable community detection method for a multiplex or multilayer lexical network is a challenge to cutting-edge research in cognitive networks. Only a few works indirectly addressed this problem so far. Kovács and colleagues \cite{kovacs2021networks} analysed the multilayer network structure of semantic, syntactic and phonological associations of several languages, including English and Hungarian. Using a modularity maximisation approach, they found that larger communities tended to reflect mostly semantic associations, whereas smaller communities encoded encyclopedic knowledge \cite{kovacs2021networks}. Interestingly, the task of grouping nodes in a multilayered network is tied to the task of detecting communities in attributed networks, where nodes also possess categorical information or features \cite{chunaev2020community}. In this direction, Citraro and Rossetti \cite{citraro2020identifying} recently tackled community detection in multilayer lexical networks by introducing the Extending to Vertex Attributes Louvain or EVA method. Their approach extends the modularity-based optimisation function \cite{blondel2008fast} to a multi-objective criterion forcing communities to be homogeneous to the features carried by the nodes and across different network layers. The authors tested the multiplex mental lexicon built in \cite{stella2018multiplex} and enriched with lexical features such as word length, valence, arousal, dominance, semantic size of denoted words and gender association (from the Glasgow dataset \cite{scott2019glasgow}). 

\subsubsection{Multiplexity highlights shortcuts between concept communities in a phonological/semantic multiplex model}

Figure \ref{fig:fig3} (A) sums up with a toy example of what community detection algorithms that are also sensitive to the features attributed to each node can reveal in multilayer lexical networks. Analysing the aggregated multiplex structure obscures the variation in feature homogeneity across layers. For instance, when forcing communities to be homogeneous for word length modularity shows a very fast decrease in the semantic layers, but remains stable in the phonological layer (Figure \ref{fig:fig3} (B)). 

Two main results emerged from the feature-rich multiplex mental lexicon (with 4000 words and 2 layers, i.e. free associations and phonological similarities) analysed by Citraro and Rossetti \cite{citraro2020identifying}: 
\begin{enumerate}
    \item Communities extracted by EVA reflect thematic contexts: Concepts can fall within different contexts according to the psycholinguistic features used to perform community detection, even keeping the network fixed. In other words, one network can give rise to many sets of communities according to the feature selected for EVA. For instance, "star" belonged to a community of words relative to astrophysics when semantic size was used as a feature for community detection on the semantic layer. On the same layer, "star" belonged to a community of words relative to "shining" when arousal was used instead (see also Figure \ref{fig:fig3} C). Such thematic coherence and context swapping were not observed in the phonological layer. These findings indicate an important interplay between semantic features (e.g. "being astrophysics objects", "shining") and semantic layers in multiplex networks, which should be treated differently from phonological similarities. While it is intuitively expected for semantic features to influence thematic coherence more prominently on semantic rather than phonological networks \cite{aitchison2012words}, these quantitative models open the way to mapping and exploring the interplay between psycholinguistic norms and network structure (see also \cite{citraro2022feature});
    \item As illustrated within Figure \ref{fig:fig3} C, semantic links provide shortcuts between different clusters of phonologically similar words that would otherwise be at a greater network distance. This happens because the homogeneity of the same feature (e.g. arousal) over communities spanning different layers changes significantly across layers. Although less thematically cohesive, the multiplex network structure provides shortcuts that make single-layer communities overlap with each other. Recent investigations indicated that semantic and phonological connections are systematically better than random links in decreasing the average network distance between words \cite{levy2021unveiling}. Hence the observed overlap in communities from different layers might be due to a potential cognitive benefit, worthy of further research.
\end{enumerate}

\subsection{Finding hidden interactions between two or more layers: Mediation, suppression and other layer-interaction mechanisms}

Given that multiplex networks encode different aspects of the mental lexicon in different layers, one might ask: “How similar are layers of a given multiplex network?”. The similarity between layers of multiplex networks has so far been addressed in pairwise comparisons, using spectral and information-theoretic distance metrics \cite{vishwanathan2010graph,schieber2017quantification,de2015structural,santoro2020algorithmic,de2016spectral,mheich2020brain,hartle2020network}. 

\begin{figure}
    \centering
    \includegraphics[width=12cm]{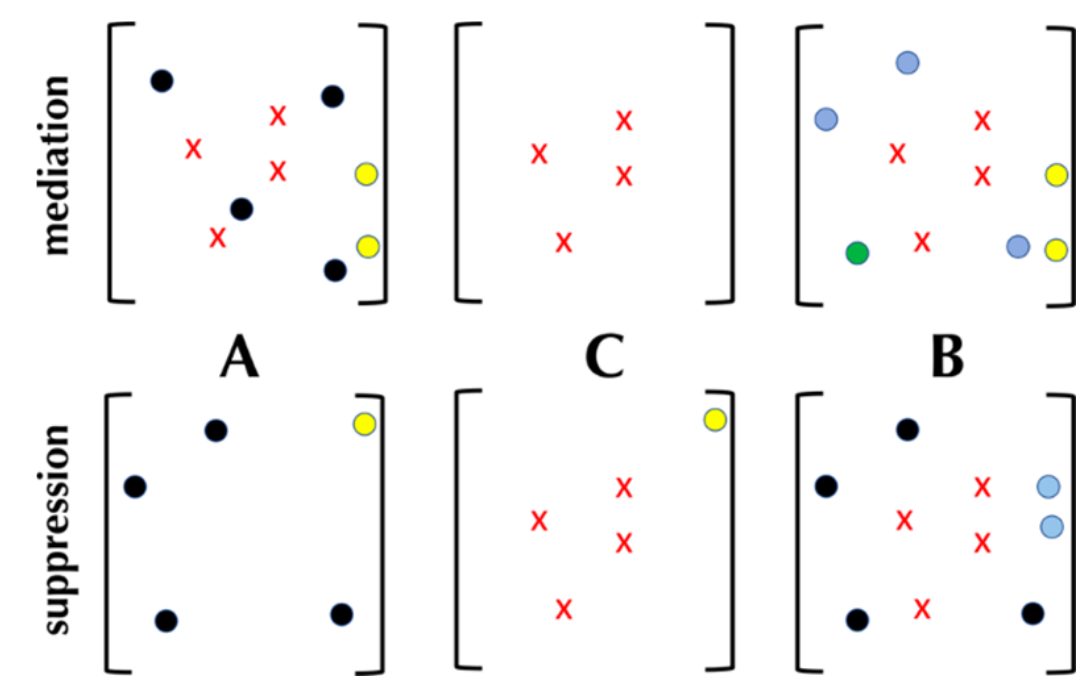}
    \caption{Schematic representations of triplets of adjacency matrices, where only the position of the edges (ones in the matrix) are highlighted as symbols (dots of different colours, crosses). Symbols are chosen to schematically highlight how mediation and suppression operate. In the mediation example - top row -  Network C mediates a relationship between A and B by sharing many links with both of them (red crosses). Most other links in A and B are different (blue and black dots), only a small fraction of them is shared directly by A and B (yellow dots). In the suppression example - bottom row - B shares many links with both A (black dots) and C (red crosses), however the links that are common to A and C (yellow dots) are suppressed in B. B contains also other links (blue dots). Reproduced from \cite{lacasa2021beyond}, copyright the authors.}
    \label{fig:fig4}
\end{figure}

Akin to what happens with pairwise correlations (and to address analogous issues related to the disambiguation of directed and mediated influences), looking at mediation and suppression is a convenient solution \cite{lacasa2021beyond}. Consider the example of a free association network (words bring each other to mind \cite{de2013better}) and of a phonological network (words sound similar \cite{vitevitch2008can}). These different layers might display a small level of correlation since words sounding similar to each other tend also to be recalled together in free association tasks (as measured in \cite{stella2017multiplex} with link overlap). However, this similarity might be due to directional relationships: Sound similarity might likely give rise to a memory recall pattern \cite{vitevitch2021cognitive}. Alternatively, this influence might not be directly evident. Analogous to latent variables in psychometrics \cite{golino2022modeling}, there might be a hidden network layer of conceptual associations that either confounds or mediates the relationship between phonological similarities and free associations or other pairs of layers in the mental lexicon.

Mediation analysis is well-established in some fields like psychometrics \cite{muthen2004latent}, but relatively unexplored in others like cognitive neuroscience \cite{parola2020pragmatics} and cognitive networks \cite{siew2019cognitive}. A recent study \cite{lacasa2021beyond} introduces a framework to quantify mediation and suppression between networks. In the case of mediation, and referring to Figure \ref{fig:fig4}, A and B are both dependent on C, such that if there is a link in C, then there is a link in A and B. For suppression, B depends on the interaction of A and C such that, an edge occurs in B with a certain probability if it appears in A but not in C, or if it appears in C but not in A.

The applications so far involved social networks (online, professional, personal interactions), and mesoscale connectomes in the complete \textit{C. elegans} nervous system \cite{lacasa2021beyond}. In lexical networks this approach could address questions such as: "Are semantic associations more likely to occur if there is semantic overlap but not phonological similarity?" If we consider factual and metacognitive layers in a knowledge network, this approach could be combined with the multiplex representation introduced by Vukić and colleagues to identify mediation and suppression mechanisms in processing domain knowledge \cite{vukic2020structural}.
Both the above examples could not be investigated with single-layer networks, highlighting the importance of using multilayer networks and mediation and suppression techniques to detect latent relationships in data due to layer-interaction mechanisms.

A concrete example of the relevance of layer-interaction mechanisms in exploring cognition comes from recent multilayer investigations of the issue of lexical access \cite{levy2021unveiling}. Classic linguistic theories assume that in order to comprehend or produce meaningful linguistic output, one needs to access and retrieve information from their mental lexicon, a process known as lexical access \cite{aitchison2012words}. Lexical access involves multiple processes of representation, in particular, a semantic word-meaning process and a phonological wordform mapping process, that allow access and retrieval from the mental lexicon \cite{dell2014word,dell1992stages}. However, whether the relation between these two processes is serial, parallel, or interactive is still debated \cite{dell2014word,nadeau2012neural}. The modular account argues for a detailed process between two discrete modular processes of lexical access. According to this account, during lexical access of a linguistic input, phonological processing takes place only after semantic processing is completed. The cascading account argues for a more relaxed modular account. According to this model, phonological processing can initiate before semantic processing is complete. Finally, the interactive model theorises that lexical access involves an interactive spread of information across a phonological layer and a semantic layer that can influence each other \cite{dell1997lexical,dell2014word}. This model argues that both layers are structured as a network and that information spreads across these two networks, related to the organisation of concepts across both layers and to the strength of links that connect them. However, the specific model of lexical access is still debate.

\subsubsection{Layer-interaction mechanisms in a phonological/semantic multiplex network}

\begin{figure}
    \centering
    \includegraphics[width=12cm]{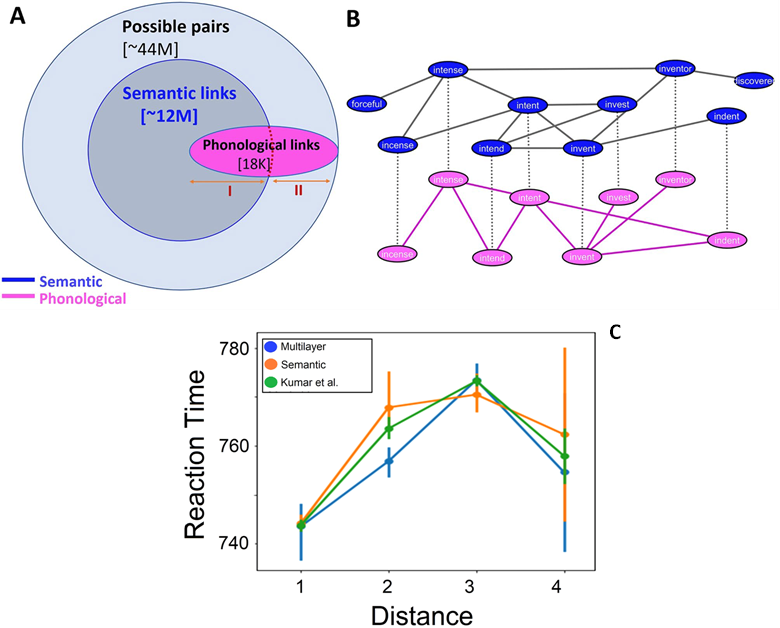}
    \caption{Top: Multiplex network of semantic and phonological layers constructed by Levy et al. (2021). (A) Illustration of the overlap across the (weighted) semantic layer and the (unweighted) phonological layer. (B) Illustration of the multiplex network with nodes and links across both semantic and phonological layers.Bottom: The multilayer network is more quickly accessible than the semantic network (C). Reaction times (RTs) for relatedness judgments in the Kumar, Balota, and Steyvers (2020) network (Green), Levy et al. (2021) semantic network (Orange) and the Levy et al. (2021) multilayer network (Blue). Error bars for the Levy et al. (2021) networks represent standard deviations of the average RT. The error bar of the Kumar, Balota, and Steyvers (2020) network represents the standard deviation of averages of subsets of RTs.}
    \label{fig:fig5}
\end{figure}

In a recent study, Levy and colleagues \cite{levy2021unveiling} applied a cognitive multilayer network analysis to directly analyse and quantify the relation between phonological and semantic networks, motivated by the interactive model of language processing \cite{dell2014word}. To do so, the authors constructed a large-scale multilayer network comprised of empirical phonological and semantic layers (see Figure \ref{fig:fig5}), for a large-scale network of about 9000 words \cite{de2019small}. The authors then conducted the following analyses: First, they examined the similarity between the two layers by measuring their link overlap. Next, they measured the effect of adding non-overlapping links from one layer to the other. Finally, Levy et al. examined the potential benefit of combining both layers as a multilayer network on lexical access, by measuring the networks’ average distances of the single layers versus the multiplex \cite{levy2021unveiling}.

Overall, the authors found that overlapping links across both layers are highly similar, that adding non-overlapping links from one layer to the other leads to reduced distances between concepts. Furthermore, a multilayer network representation has the shortest average shortest paths, indicating that it leads to the most efficient linguistic processing. Thus the authors argue that the interaction between these two layers might be crucial for allowing more efficient lexical access, by reducing path distances between nodes in a cognitive multilayer network \cite{levy2021unveiling}. For example, in the phonological layer, the cue words intend and invest had a path distance of three (intend → intent → invent → invest). However, in the semantic layer these cue words intend and invest are directly connected, see Figure \ref{fig:fig5} (B). Thus, in a multilayer phonological-semantic network, the distance between intend and invest is much shorter than in a phonological only network, enhancing the lexicon’s efficiency in lexical access even in potential impairments \cite{castro2019multiplex,castro2020quantifying}.

To demonstrate the validity of their claims regarding the heightened efficiency of the cognitive multilayer network representation in lexical access, Levy and colleagues re-analysed empirical data collected by Kumar and colleagues \cite{kumar2020distant}. The latter estimated a large-scale semantic network based on the University of South Florida Free Association Norms, generated to a list of 5000 cues \cite{nelson2004university}. Kumar and colleagues aimed to replicate and extend a previous study by Kenett and colleagues on the impact of semantic network distance on behaviour \cite{kenett2017semantic}. To do so, Kumar and colleagues had participants make relatedness judgements to pairs of cue words that varied in the semantic distance between the words based on the path length in their semantic network for those two words \cite{kumar2020distant}. Levy and colleagues examined the data collected by \cite{kumar2020distant} in the following way: First, they identified links from the Kumar et al. semantic network that corresponded with their semantic network. Then, the authors compare RT collected by Kumar et al. to various semantic distances (1-4 steps) of the Kumar et al. semantic network, the Levy et al. semantic network, and the Levy et al. multilayer network. The authors show that overall, similar distance effects on RT were consistent across all three examined networks \cite{levy2021unveiling}. However, the multilayer network predicted shorter RT responses than either semantic network highlighting its heightened efficiency in lexical access, see Figure \ref{fig:fig5} (C).

Overall, the multilayer approach by Levy and colleagues demonstrates the strength of applying a cognitive multilayer network analysis to examine classic cognitive theories, such as on the nature of lexical access \cite{levy2021unveiling}. It also demonstrates the feasibility of combining computational modelling with empirical research to advance cognitive research.

\section{Discussion, limitations and future directions}

Recent work using multilayered networks in cognitive science has revealed key insights that would not have been observed using single-layer networks. This work has examined cognitive processing in healthy \cite{stella2019viability,stella2018distance,stella2017multiplex,stella2018multiplex,levy2021unveiling} and clinical populations \cite{castro2020quantifying,castro2019multiplex,stella2017multiplex}, revealed clusters and communities reflecting different contexts and meanings of individual concepts \cite{citraro2020identifying,citraro2022feature,kovacs2021networks,vukic2020structural}, and discovered latent mediation/suppression interactions between different aspects of knowledge \cite{marinazzo2022information,vukic2020structural}. These findings illustrate the potential for multilayer networks to advance the cognitive sciences using a quantitative, interpretable, and human-centric framework. Multilayer networks give structure to the representations found in interactive layers of the mental lexicon (quantitative \cite{de2022multilayer}). This structure can be interpreted using various network measures, such as network distance and concept relatedness (interpretability, e.g.  \cite{kumar2020distant,kenett2017semantic}), and may account for the complexity of the human mind (human-centric \cite{stella2019modelling}). 

Although the use of multilayer networks has much potential, this approach also has some limitations, which naturally lead to crucial directions for future research. For example, in most cases (including the examples presented here) links are defined between pairs of nodes. In some cases, it may be more useful to create a set of nodes instead of simply pairs of nodes to form a hypergraph. For example, several actors feature in scenes in movies without just co-occurring with each other, or more than two words in a sentence might modify its meaning. Hypergraphs are being increasingly studied and applied \cite{battiston2020networks} to account for simultaneous interactions between more than two entities at once. From a cognitive perspective, hypergraphs could be structured across multilayer/multiplex structures either by building links through information-theoretic measures \cite{marinazzo2022information} or by considering other interaction patterns between concepts (e.g. phonological similarities between words sharing the same skeleton of vowels and consonants \cite{gruenenfelder2009lexical}). In either case, future research using the hypergraph approach could potentially reveal higher-order behaviours that are not observable with pairwise relationships, perhaps identifying communities of concepts reflecting specific semantic fields \cite{gerow2014modular} or contexts of usage \cite{citraro2022feature}.

Multilayer networks contain several network layers and thus correspond to an increased chance of mistakes in assessing whether two concepts should be connected or not, such as whether two concepts are syntactically related in speech \cite{morgan2021natural,parola2022speech} or in text \cite{semeraro2022emotional}. Mechanisms for link prediction or noise correction should thus be applied to next-generation multilayer models of the mental lexicon. Bayesian inference can identify errors in a given network layer even in the presence of unknown and heterogenous uncertainty \cite{peixoto2018reconstructing}. The formalism proposed by Peixoto checks for the presence of connections between different clusters of nodes, using structured generative network models to infer the presence/faulty presence of individual links even without direct error estimates. This approach could filter layers of free associations \cite{de2013better} or other types of semantic/syntactic datasets \cite{quispe2021using}, lessening the impact of noise over the multilayer structure.

Single-layer and multilayer networks are useful constructs to make sense of complex and multivariate systems \cite{newman2018networks}, but they remain modelling proxies. The mental lexicon might not look like a network in some specific instances, as recently discussed by Hills and Kenett \cite{hills2022mind}. Alternative modelling approaches should thus be pursued in parallel with multilayer networks, leading to next-generation studies where multiple models are compared or used together. Word embeddings \cite{kennington2015simple,kumar2020distant,litovsky2022semantic,parola2022speech} represent promising alternative modelling approaches, giving more emphasis to the vector-like nature of features associated with individual concepts. Frameworks encompassing vectors and multilayer networks to model the mental lexicon, like the FERMULEX approach by Citraro and colleagues \cite{citraro2022feature}, represents an interesting attempt to capture the complexity of mental representations. 

Building network models that minimise redundant features while maximising informativeness (e.g. prediction power about word norms \cite{scott2019glasgow}) under potential uncertainty is becoming increasingly relevant in quantitative psychology, especially network psychometrics \cite{golino2022modeling,christensen2020unique}. A current limitation of cognitive multilayer networks is the selection of which layers to include and which to discard when building a representation of the mental lexicon \cite{stella2017multiplex}. Adding more network layers can provide more information about connectivity patterns between concepts but, at the same time, it can add unnecessary redundancy \cite{santoro2020algorithmic}. This limitation can be addressed in two ways. First, representations of the mental lexicon should include only the layers that are relevant for a specific task. For example, to model the process of reading the interplay between phonology and orthography requires both layers \cite{siew2019phonographic}. Second, once relevant layers have been selected, additional tools from information theory can quantify the amount of redundant information embedded in a given combination of layers. Structural reducibility analysis \cite{de2015structural} and compressibility \cite{santoro2020algorithmic} can identify the best combination of network layers maximising information gain compared to a baseline model where all layers are aggregated together. Information gain could be implemented in different ways \cite{santoro2020algorithmic}. In multiplex networks, the maximisation of the Von Neumann entropy was shown to successfully identify those layers providing the most information about node connectivity \cite{de2015structural}. The experimenter should thus check whether a preliminary representation of the mental lexicon could be further aggregated or compressed via entropy maximisation, which would indicate the presence of redundant layers to be aggregated with each other to limit the number of layers to be considered. These approaches unveiled that many social and technological multilayer networks exhibited moderate redundancy \cite{santoro2020algorithmic,de2015structural,de2016spectral} and could be further reduced/compressed in structure. This was not the case for all multiplex lexical networks reviewed here, which showed how semantic, syntactic, and phonological aspects of words captured very different, and thus irreducible, patterns of connectivity \cite{stella2017multiplex,stella2018multiplex,stella2019viability}. Future research should combine informed designs and information-theoretic tools to better select appropriate layers for and reduce redundancy in a given multilayer model of the mental lexicon.

Importantly, networks are not only being used to understand the complexity of the human mind \cite{hills2022mind,vitevitch2008can,stella2018cohort}, but are also being employed to understand the complexity of the human brain \cite{bullmore2011brain,betzel2017multi,aerts2016brain,amico2021toward}. These single-layer networks of the brain may connect brain regions that are physically connected or brain regions that are active at the same time. An ambitious goal for future research is to use the multilayer network approach to connect the cognitive network layer to the brain network layer to finally bridge the intangible mind and the physical brain \cite{vitevitch2021cognitive,zaharchuk2021multilayer}. At present, it is not clear how many network layers would be needed to accomplish this, or what those intermediate network layers might represent. It is also not clear if the spread or diffusion of activation that is commonly used to model cognitive processing in cognitive network models \cite{vitevitch2021cognitive,litovsky2022semantic} is an appropriate mechanism to model the processes that occur at other network layers. Connecting the mind and the brain using multilayer networks may seem like an elusive goal, but we need only look to the Internet for an existence-proof of a physical network (i.e., the fiber-optic cables that envelope the world) bridging to the intangible social networks that emerge on software platforms like Facebook, Twitter, etc. (whose information is transmitted across those fiber-optic cables). Future research bridging cognitive and brain networks within multilayer, feature-rich frameworks might contribute to building a quantitative understanding of how the brain stores conceptual representations of words, which represents an intriguing brain/mind puzzle \cite{poeppel2022brainmind}.  

\section{Conclusions}

Cognitive multilayer networks can map multiple types of cognitive information at once. Their quantitative framework can thus model how different types of associations might co-exist within the mental lexicon and influence cognitive processing. This review has highlighted several pioneering studies unearthing mechanisms of psychological phenomena that could not be observed in single-layer cognitive networks. The phenomena that were only unveiled by the combination of multiple layers of associative knowledge included: (i) multiplex viability as a booster of lexical search and processing in people with lower creativity, shielding words from degraded production in people with aphasia, (ii) multilayer community detection as a way to highlight thematic clusters of concepts shaped by psycholinguistic norms and linked by multilayer shortcuts, and (iii) layer-layer correlations as interactive mechanisms between phonological and semantic similarities in lexical processing. In addition to describing the novel quantitative perspectives where multilayer networks can shed light on knowledge representations in the mental lexicon and in potential brain/mind models, we have discussed key limitations and promising directions for future research. The formalism covered in this review thus opens the way to next-generation quantitative frameworks of cognition able to model multivariate psychological data.

\backmatter

\section*{Declarations}

The authors declare no conflict of interest.

\bibliography{sn-article}


\end{document}